\documentclass[10pt,twocolumn,letterpaper]{article}

\usepackage{iccv}
\usepackage{times}
\usepackage{epsfig}
\usepackage{graphicx}
\usepackage{amsmath}
\usepackage{amssymb}
\usepackage{booktabs}
\usepackage{multirow}
\usepackage{subcaption}
\usepackage{ulem}
\usepackage[ruled,vlined,linesnumbered]{algorithm2e}
\usepackage[table,xcdraw]{xcolor}

\usepackage[breaklinks=true,bookmarks=false]{hyperref}

\iccvfinalcopy 

\def\iccvPaperID{5669} 

\ificcvfinal\fi

\begin{document}

\title{Improving Generalization in Visual Reinforcement Learning via\\ Conflict-aware Gradient Agreement Augmentation}
\author{
Siao Liu \and Zhaoyu Chen \and Yang Liu \and Yuzheng Wang \and Dingkang Yang \and Zhile Zhao \and Ziqing Zhou \and Xie Yi \and Wei Li \and Wenqiang Zhang \and Zhongxue Gan \and \\
Academy for Engineering \& Technology, Fudan University\\
{\tt\small \{saliu20, zhaoyuchen20, yang\_liu20, yzwang20, dkyang20, fd\_liwei, wqzhang, ganzhongxue\}@fudan.edu.cn} \and {\tt\small \{zhilezhao21, ziqingzhou21, yixie22\}@m.fudan.edu.cn}
}

\maketitle
\ificcvfinal\thispagestyle{empty}\fi
\begin{abstract}
Learning a policy with great generalization to unseen environments remains challenging but critical in visual reinforcement learning. Despite the success of augmentation combination in the supervised learning generalization, naively applying it to visual RL algorithms may damage the training efficiency, suffering from serve performance degradation. In this paper, we first conduct qualitative analysis and illuminate the main causes: (i) high-variance gradient magnitudes and (ii) gradient conflicts existed in various augmentation methods. To alleviate these issues, we propose a general policy gradient optimization framework, named Conflict-aware Gradient Agreement Augmentation (CG2A), and better integrate augmentation combination into visual RL algorithms to address the generalization bias. In particular, CG2A develops a Gradient Agreement Solver to adaptively balance the varying gradient magnitudes, and introduces a Soft Gradient Surgery strategy to alleviate the gradient conflicts. Extensive experiments demonstrate that CG2A significantly improves the generalization performance and sample efficiency of visual RL algorithms.
\end{abstract}

\section{Introduction}

With the development of deep learning in various tasks~\cite{g1,g2,g3,g4,g5,g6,g11,g12,g13,g14}, visual Reinforcement Learning (RL) has achieved impressive success in various fields such as robotic control~\cite{rlrobot}, autonomous driving~\cite{rlvehicle}, and game-playing~\cite{rlgame}. Previous works usually formulate it as a Partially Observable Markov Decision Process (POMDP)~\cite{pomdp}, and the agent receives high-dimensional image observations as inputs. As depicted in~\cite{soda,svea}, visual RL generalization refers to the ability of a pretrained RL agent to perform well in unseen environments. Due to the dynamic nature of the real world, even minor perturbations in the environment can result in significant semantic shifts in the visual observations, which makes visual RL generalization challenging.

To improve generalization performance, data augmentation~\cite{multiple} is a widely adopted technique in reinforcement learning. Numerous studies~\cite{rad,pad} utilize data augmentation methods to generate synthetic data and diversify the training environments, yielding considerable performance improvements. However, recent methods~\cite{svea,sgqn,pieg} mostly select a single augmentation technique to improve the generalization capability, resulting in a poor performance in the environments with observations varying far from the augmented images. For instance, ColorJitter~\cite{colorjitter} is the preferred choice for addressing color variations, but agents trained with such augmentation still hard to cope with intricate texture patterns. In other words, the generalization ability heavily relies on the selection of specific data augmentation technique, which is so-called generalization bias.

Compared to single data augmentation, Augmentation Combination (AC)~\cite{moco} integrates multiple data augmentation methods to enhance the diversity of augmentations and alleviate the generalization bias, which is a more promising pre-processing solution. Unfortunately, there is a dilemma in incorporating AC into visual RL. Although data augmentation combination can effectively improve generalization capability in the supervised visual tasks, RL algorithms are quite sensitive to excessive variations, resulting in performance degradation and training sample inefficiency. Therefore, it is necessary to rethink why visual RL algorithms cannot benefit from AC as much as supervised learning.

From the perspective of gradient optimization, we conduct numerous qualitative analysis to illustrate the causes of performance degradation and training collapse that occur when employing augmentation combinations during training. There are two primary reasons for this phenomenon: (\textbf{i}) the utilization of diverse data augmentations leads to high gradient magnitude variations, resulting in biased generalization; (\textbf{ii}) the gradient conflicts\footnote{The gradient conflicts mean the gradient directions point away from each other, \eg, appears a negative cosine similarity.}~\cite{pcgrad} existed cross multiple augmentation methods hinder the policy optimization. To balance the gradients with high-variance magnitudes, one effective approach is to customize the weights of the loss terms with manually defined hyper-parameters~\cite{svea}. However, hyper-parameter tuning relies heavily on expert knowledge, which can be inflexible and computationally expensive when dealing with multiple data augmentations. Besides, ~\cite{andmask} indicates that the widely employed average-based gradient update strategies tend to converge towards the speed-greedy direction, and are ill-posed to effectively handle complex gradient conflicts, leading to local optima and a decrease in sampling efficiency.

To address these issues, we propose a general policy gradient optimization framework, named Conflict-aware Gradient Agreement Augmentation (CG2A), to integrate augmentation combination into the RL framework and improve its generalization performance. Specifically, the CG2A contains two key components: an adaptive weight assigner called Gradient Agreement Solver (GAS) and a conflict-aware gradient update strategy Soft Gradient Surgery (SGS). To effectively harmonize high-variance magnitudes gradients, we formulate the hyper-parameter tune as a second-order multi-objective optimization problem and use the GAS to obtain a proximal approximate solution with minimal computational cost. Moreover, according to~\cite{andmask}, although gradient conflicts slow down convergence speed,  these conflicting gradient components may contain more semantic-irrelevant information that can improve invariant learning consistency. Motivated by this hypothesis, we propose SGS to improve the gradient update process, which preserves a small amount of conflicting gradient components to strike a balance between convergence speed and generalization performance. To validate the effectiveness of CG2A, we conduct extensive experiments on DMControl Generalization Benchmark (DMC-GB) and some robotic manipulation tasks. In summary, our contribution encompasses three main manifolds:
\begin{itemize}
    \item We point out the generalization bias induced by single data augmentation and illustrate the primary causes for performance degradation when naively applying augmentation combination in RL algorithms. 
    \item We propose a general policy gradient optimization framework named Conflict-aware Gradient Agreement Augmentation (CG2A), to efficiently integrate data augmentation combinations into the RL algorithms and significantly improve the generalization performance in various environments.
    \item We devise a Gradient Agreement Solver (GAS) to harmonize multiple gradients with high-variance magnitudes, and propose a Soft Gradient Surgery (SGS) strategy to alleviate the gradient conflicts existed in various data augmentations.
    \item Compared to previous state-of-the-art methods, CG2A achieves competitive generalization performance and significantly improves sample efficiency.
\end{itemize}

\section{Related Work}

\subsection{Data Augmentation in Visual RL}

Benefiting from the development in the field of computer vision~\cite{mixup,g7,g8,g9,cutout}, data augmentation is widely used in the visual RL~\cite{rad,svea,soda}. As noted by Kirk \etal ~\cite{zero-shot-survey}, DA force the agents to learning an invariance knowledge through regularising models to have same output or inherent representation for different augmented images. Kostrikov \etal~\cite{drq} adopt simple pixel-level transformations to perturb image observations and regularize the value function and policy, which significantly boost the sample efficiency. Inspired by MixUp~\cite{mixup}, Wang \etal~\cite{mixreg} propose to train agents with a mixture of observations and impose linearity constraints to improve the generalization capability. Meanwhile, Raileanu \etal~\cite{DrAC} propose a principle to automatically select an effective augmentation from a set of data augmentations for RL tasks. More recently, task-aware data augmentation with Lipschitz constant is devoloped~\cite{tlda}, which maintain the sample efficiency and alleviate instability caused by the aggressive data augmentations.  Unlike prior work, our method expect to explore the utilization of augmentation combinations in visual RL to improve generalization capability in various unseen environments instead of a single manually or automatically data augmentation.

\subsection{Generalization in Visual RL}
Numerous studies~\cite{drq,soda,svea,pieg} attempt to enhance the generalization capability of agents through various approaches, such as data augmentation~\cite{rad,tlda,drq}, domain randomization~\cite{domainrand}, and self-learning based methods~\cite{soda}.
Hansen \etal~\cite{soda} build a BYOL~\cite{byol}-like architecture and use an auxiliary loss to foster the representations to be invariant with the irrelevant perturbations. Hansen \etal~\cite{svea} introduce a regularization term for the Q-function which reduces variance implicitly by linear combining the estimated Q-value between unaugmented and augmented data.
Bertoin \etal~\cite{sgqn} integrate saliency maps into the RL~\cite{sac} architecture, enabling the agent to guide its focus towards the most salient aspects of the observation images during the decision-making process.
Yuan \etal~\cite{pieg} utilize a pre-trained model and extract generalizable representations from the early layers of the encoder for enhancing the generalization performance. 
Unlike previous work, we utilize data augmentation combination to eliminate the generalization bias and develop an effective optimize framework to avoid the performance collapse during the training stage.
\section{Preliminaries}

\noindent \textbf{Reinforcement Learning.} Considering that the agent cannot directly observe the underlying state of environment from the given images~\cite{po}, visual RL~\cite{vrl} formulates the interaction between the agent and its environment as a discrete-time Partially Observable Markov Decision Process (POMDP). Formally, a POMDP can be defined as a 6-tuple $\langle \mathcal{O}, \mathcal{S}, \mathcal{A}, \mathcal{P}, r, \gamma \rangle$, where $\mathcal{O}$ is the high-dimensional observation space, $\mathcal{S}$ is the state space, $\mathcal{A}$ is the action space, $\mathcal{P}(s_{t+1}|s_t,a_t)$ is the conditional transition function between states, $r:\mathcal{S} \times \mathcal{A} \to \mathbb{R}$ is the reward function, and the $\gamma \in \left [ 0,1 \right )$ is the discount factor~\cite{pomdp}.

\noindent\textbf{Generalization Definition.} Given a set of POMDPs $\mathbb{M}=\{\mathcal{M}_1,\mathcal{M}_2,\ldots,\mathcal{M}_n\}$, each POMDP $\mathcal{M}_i$ has its own observation space $\mathcal{O}_i$, but shares a common underlying state space $\mathcal{S}$ and dynamic conditional transition function $\mathcal{P}$. Our objective is to learn a general policy that can generalize to unseen environments in a zero-shot manner. Specifically, we have access to only one specific POMDP $\mathcal{M}_i \in \mathbb{M}$ and utilize it to learn a general policy $\pi^{*}$. The policy $\pi^{*}$ is expected to alleviate the dependency on the individual observation space $\mathcal{O}_i$ and explore the inherent state structure to perform well over the whole set of $\mathbb{M}$.

\begin{figure}[tbp]
\centering

\begin{subfigure}{1\linewidth}
    \includegraphics[width=0.97\linewidth]{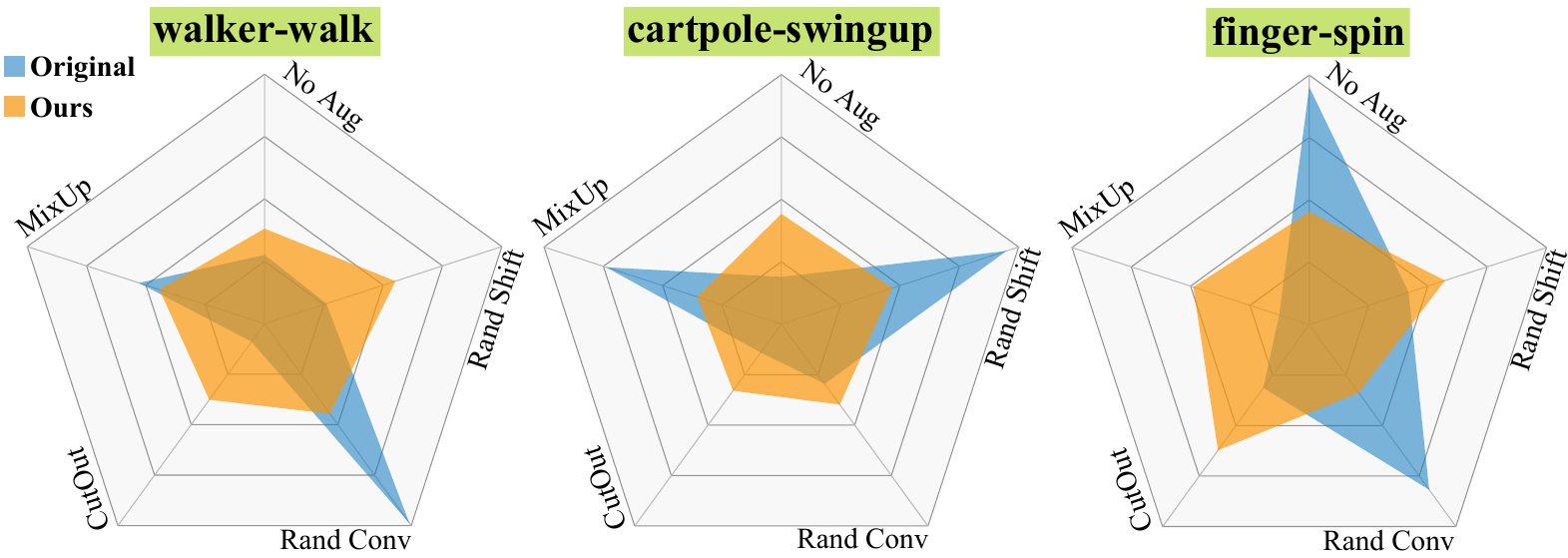}
    \caption{\small(a) \textit{Visualization of normalized gradients' magnitude associated with various augmentations in AC.} For each task, the optimization process can be dominated by specific gradients. Our method effectively suppresses these aggressive gradients.}
\end{subfigure}

\begin{subfigure}{\linewidth}
    \includegraphics[width=\linewidth]{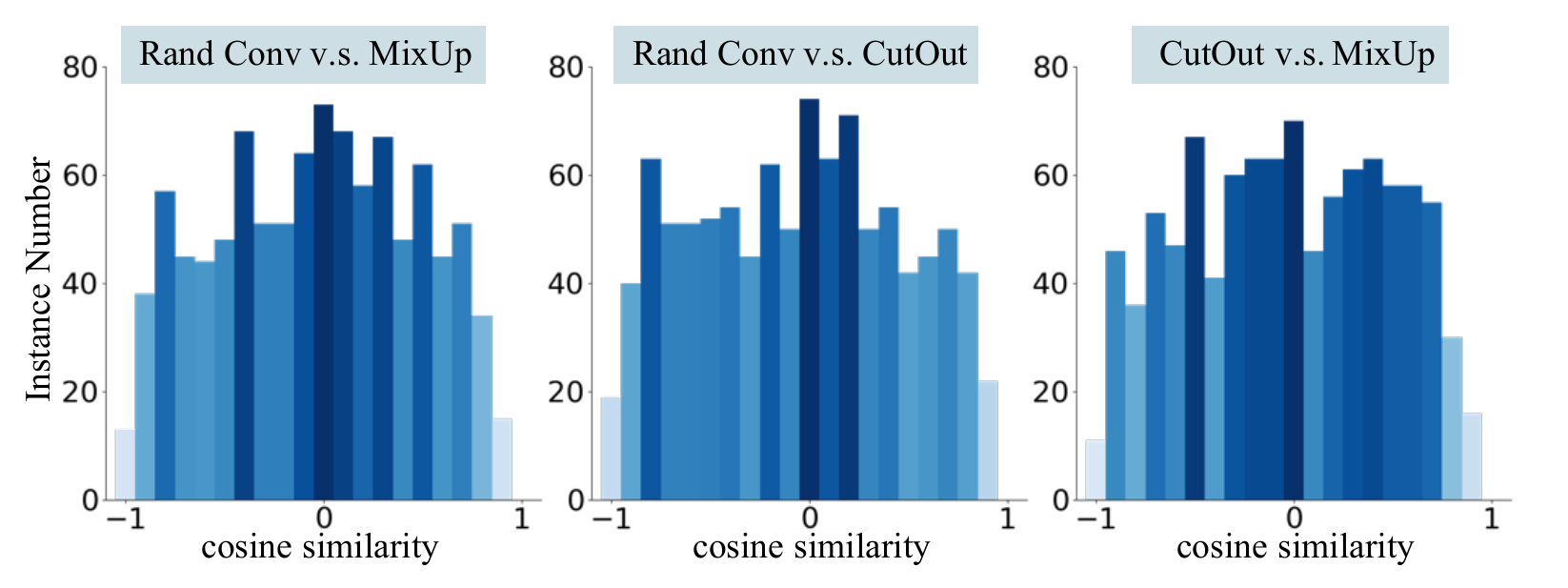}
    \caption{\small(b) \textit{Statistics of cosine similarity between paired gradients for various augmented data.} There are a lot of paired gradients with a negative direction, which misleads gradient optimization.}
\end{subfigure}

\caption{The qualitative analysis of gradient magnitudes and gradient directions for the gradient vectors associated with data equipped with various augmentations.}
\label{analyse}
\vspace{-16pt}
\end{figure}

\noindent \textbf{Optimization Pitfall with Augmentation Combination.}
Here, we elucidate the main reasons for limited training performance when data augmentation combinations are naively applied in policy gradient optimization.
Specifically, we choose a random initialized SAC~\cite{sac} agent as a base model and collect multiple data augmentation technologies, including random shift~\cite{drq}, random convolution~\cite{rad}, CutOut~\cite{cutout}, and MixUp~\cite{mixup}, to form a data combination. Then, a SVEA~\cite{svea} architecture agent is adopted to integrate the data combination and calculate corresponding critic loss items associated with each data augmentation and obtains the corresponding gradient via backpropagation. Based on such framework, we conduct various toy experiments on DMC~\cite{dmc} suite and analyse the composition of the gradient from the perspective of the magnitude and direction. There are two primary reasons. \textbf{(i)} \textit{High-variance gradient magnitudes}: we collect the normalized gradients associated with various augmented data over 5000 times for a same agent and compute the arithmetic mean as the  gradient magnitude representation for such augmentation. As shown in Figure~\ref{analyse}(a), the empirical results show that all agents exhibit high variance in gradient magnitude over the three tasks and that certain gradients prevail in magnitude, implying that the corresponding data augmentations dominate the policy optimization and lead to significant generalization bias. \textbf{(ii)} \textit{Gradient conflicts existed in multiple DAs}: we sample 1000 image observations from an unseen environments and record the mean of gradient cosine similarity over 5 times cross paired augmented data. In our context, we define two gradients as conflicting if they have negative cosine similarity, indicating that they are far apart from each other. According to~\cite{pcgrad}, if two gradients conflict with each other, the agent would struggle to find a balance between various optimization. Figure~\ref{analyse}(b) indicates that gradient conflicts are a prevalent issue among different data augmentations, leading to slower convergence and performance degradation.





\begin{figure*}[t]
    \begin{center}
    \includegraphics[width=0.9\linewidth]{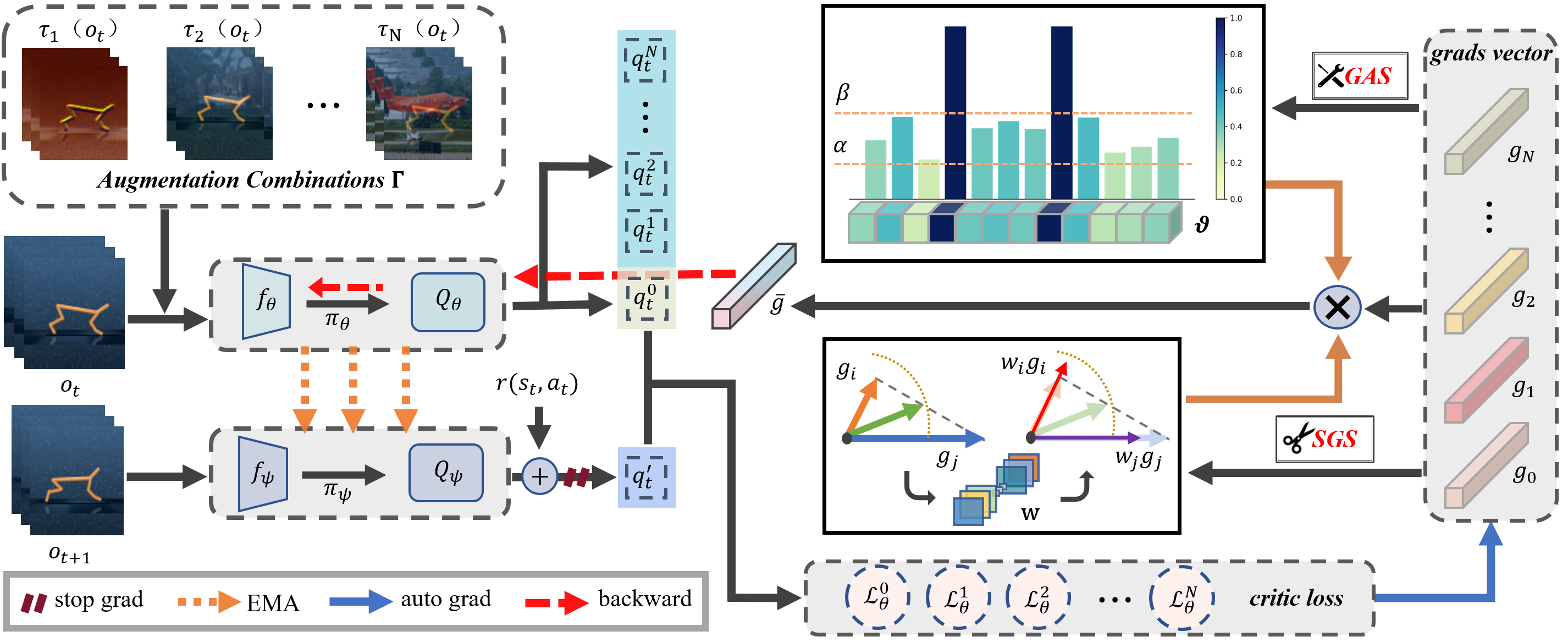}
    \end{center}
    \vspace{-8pt}
    \caption{\textbf{Overview.} 
    We input the original observation and augmented data and  obtain the estimated Q-values $q_t^*$ to calculate the loss items and the corresponding gradients $g_i$.
   Given the gradient vectors, CG2A utilizes SGS and GAS to calculate the weight coefficients $\textbf{w}$ and gradient masks $\vartheta$ respectively and update the policy parameters with generated $\hat{g}$.}
    \label{overview}
    \vspace{-8pt}
\end{figure*}

\section{Method}
In this section, we propose a general policy gradient optimization framework, named Conflict-aware Gradient Agreement Augmentation, to address these pitfalls from the perspective of gradient magnitude and direction.
\subsection{Overview}
 Soft-critic-actor~\cite{sac} (SAC) is implemented as the basic visual reinforcement learning algorithm. Briefly, SAC simultaneously learns a stochastic policy and a Q-function to maximize the discounted rewards, while incorporating entropy regularization to encourage the policy to explore diverse actions. Sampling a mini-batch of transitions $\zeta = \{o_{t}, a_{t}, r(s_t,a_t),o_{t+1}\}$ from the replay buffer $\mathcal{B}$, the critic loss function can be expressed as follows:
\begin{equation}
    \label{loss_sac}
    \mathcal{L}_{\theta} = \sum_{\zeta\in \mathcal{B} }||Q^*_\theta(o_t, a_t) -  \hat{q_t}||^2,
\end{equation}
with the Temporal Difference (TD) target $\hat{q_t}$:
\begin{equation}
    \hat{q_t}=r(o_t,a_t) + \max\limits_{\hat{a}} Q_{\psi}(s_{t+1},\hat{a}).
\end{equation}
To correct the generalization bias caused by single data augmentation, we gather a collection of diverse augmentations, \eg, random shift~\cite{drq}, random conv~\cite{rad} and MixUp\cite{mixup}, to construct an augmentation combination denoted as $\Gamma =\{\tau_1, \tau_2,\ldots,\tau_N\}$, where $N$ is the number of augmentations.
As shown in the Figure~\ref{overview}, we inject the vanilla observation $o_t$ and all augmented data $\{\tau_1(o_t),\ldots,\tau_N(o_t)\}$ into the network $\pi_{\theta}$ and then obtain respective estimated state-action value $\{q_t^0,q_t^1,\ldots,q_t^N\}$. As per the recommendation in ~\cite{svea}, we employ no data augmentation on the successor observations to maintain a deterministic Q-target $\hat{q_t}$. 

To mitigate the high-variance cross various gradient magnitudes, we associate an adaptive weight coefficient $w_i$ to each critic loss term and reformulate it as follows:
\begin{equation}
    \label{loss_sacs}
    \mathcal{L}_{\theta} = \sum_{i=0}^N \sum_{\zeta\in \mathcal{B} } 
 w_i||Q^*_\theta(\tau_i(o_t), a_t) - \hat{q}_t||^2.
\end{equation}
Then, we joint the optimization of coefficient $\textbf{w}$ into the agent policy training and formulate it as a multi-objective optimization. Motivated by~\cite{implict}, we apply proximal approximation
using Taylor series and obtain an optimal solution of the weights $\textbf{w}$. To directly adjust the overall gradient magnitude, we opt to multiply the weight coefficients $w_i$ with the individual gradient vectors $g_i$ in the practical implementation. Next, we utilize the soft gradient surgery to locate the conflicting gradient components in the gradient vectors and randomly clip a certain proportion of conflict components to rectify the gradient direction. Considering the above process only involves linear or sign operations, the two modules can be computed independently and use linear multiplication to obtain the final update gradients $\overline{g}$. More details are shown in Figure~\ref{overview}.

\subsection{Gradient Agreement Solver}
\label{gas}

To avoid some dominant augmentations to misguide the policy gradient optimization, we devise gradient agreement solver to adaptively assign weight coefficients $\textbf{w}$ to all loss items, which can directly affect the gradient magnitude. Here, we aim to simultaneously optimize the policy parameters $\theta$ and find optimal coefficients $\textbf{w}$, and thus we model it as a two order multi-objective optimization process. Additionally, we incorporate $L2$ regularization into the objective function to mitigate over-fitting and promote the weights smoothing, defined as follows:
\begin{equation}
\begin{aligned}
    \label{eq4}
    \hat{\theta},\hat{\textbf{w}} &= \mathop{\arg\min}\limits_{\theta,\textbf{w}}\sum\limits_{i \in T} \mathcal{L}_{\theta}^{i}(\theta(\textbf{w})) + \lambda ||\textbf{w}||_{2}^{2} \\
     \text {s.t.}\ \theta(\textbf{w}) &= \mathop{\arg\min}\limits_{\overline{\theta}} w_i\mathcal{L}_{\theta}^i,(\overline{\theta}), ||\textbf{w}||_1 = 1,
     \end{aligned}
\end{equation}
where $\lambda$ is a regularization item. Considering that the optimal solution is not influenced by the scale of $\textbf{w}$, we normalize the magnitude of $\textbf{w}$ to one by default. A typical method of addressing second order derivative problems is the implicit differentiation~\cite{implict}. Briefly, implicit differentiation solves second-order optimization problems by differentiating the equation of the objective function, and then solving for the rate of change of the variable being optimized, which is computationally intensive.
To reduce computational cost, we assume the objective function have sufficient differentiability and then adopt the Taylor series to derive a proximal approximation. Specifically, we approximate the $i$-th loss item $\mathcal{L}_\theta^{\tau_i}$ in Eq.~\eqref{eq4} using the Taylor series as follows:
\begin{equation}
    \label{eq5}
    \mathcal{L}_{\theta}^i(\theta) \approx \mathcal{L}_{\theta}^i(\theta^t) + g_i(\theta - \theta^t).
\end{equation}
By plugging Eq.~\eqref{eq5} into Eq.~\eqref{eq4} and temporarily ignoring the $L_1$ normalization constraint, we can obtain the following objective function:
\begin{equation}
\resizebox{0.87\width}{!}{$
\begin{aligned}
 \label{eq6}
    \theta^{t+1},\hat{\textbf{w}} &= \mathop{\arg\min}\limits_{\theta,\textbf{w}}\sum\limits_{i \in T} [\mathcal{L}_{\theta}^i(\theta^t) + g_i(\theta(\textbf{w})-\theta^t)]\\
     \text { s.t. }\theta(\textbf{w}) &= \mathop{\arg\min}\limits_{\overline{\theta}} w_i [\mathcal{L}_{\theta}^i(\theta^t) + g_i(\overline{\theta} - \theta^t) ] + \frac{||\overline{\theta}-\theta^t||_{2}^{2}}{2\epsilon},
     \end{aligned}$}
\end{equation}
The closed-form solution~\cite{close} to the quadratic problem in Eq~\eqref{eq6} can be $\theta(\textbf{w}) = \theta_t -\epsilon \textbf{w}^T\textbf{g}$, which is a classical SGD update process, and we compute the derivative:
\begin{equation}
    w_i = \lambda \sum\limits_{j \in T} (g_i^Tg_j).
\end{equation}
Finally, we add the $L_1$ constraint to normalize the weight coefficients $\textbf{w}$, we obtain $w_i$ as follows:
 \begin{equation}
     w_i = \frac{\sum_{j=0}^N(g_i^Tg_j)}{\sum_{k=0}^N|\sum_{j=0}^N(g_k^Tg_j)|}. 
 \end{equation}
 Intuitively, we can regard the weight $w_i$ as a linear expansion of a dot product between the gradient vector $g_i$ and the average of all gradients.
Hence, the GAS entails assigning greater weights to loss items that exhibit well-aligned with the average of all gradients. Consequently, GAS enable to guide the policy optimize towards a direction that exhibits greater agreement among all data augmentations.
\begin{algorithm}
\normalem
\SetAlgoLined
\KwIn{Hyper-parameters $\alpha$, $\beta$; gradient set $\mathbb{G} = \{g_0,g_1,\ldots,g_N\}$} 
\KwOut{The new gradient set $\overline{\mathbb{G}}$}
\tcp{\small flatten each gradients}
\ForEach{$g_i$ in gradient set $\mathbb{G}$}
{$g_i \leftarrow \text{flatten}(g_i)$}
$M \leftarrow len(g_0)$\;
$G \leftarrow \text{concat}(g_0,g_1,\ldots,g_N)$\;
\tcp{\small obtain the elemental mask}
\For{$j\leftarrow 1$ \KwTo $M$}{
        \eIf{$\sum_{i=0}^{N} \text{sign}(g_i^j)= N+1$}{
             $\vartheta[j] \leftarrow 1$\;
        }{$\vartheta[j] \leftarrow 0$\;}}
\tcp{\small randomly clipping the conflicting gradient components}
$\gamma \leftarrow \text{randomly sample from} \textbf{ U}(\alpha,\beta)$\;
$G \leftarrow \vartheta\times G + \gamma\times \sim~\vartheta\times G$\;
\For{$i\leftarrow 0$ \KwTo $N$}
{$\overline{g_i} \leftarrow \text{reshape}(G[i])$}
\textbf{return} $\{\overline{g_0},\overline{g_1},\ldots,\overline{g_N}\}$
\caption{Soft Gradient Surgery.}
\label{sgs}
\end{algorithm}

\begin{table*}[]
\begin{center}
\renewcommand\arraystretch{1.2}
\resizebox{\textwidth}{!}{
\begin{tabular}{@{}ccccccccccccc@{}}
\toprule                       & DMCGB~\cite{soda}               & SAC~\cite{sac}         & DrQ~\cite{drq}         & DrQv2~\cite{drqv2}      & RAD~\cite{drqv2}         & PAD~\cite{pad}        & SODA~\cite{soda}        & SVEA~\cite{svea}       & TLDA~\cite{tlda}       & PIE-G~\cite{pieg}      & SGQN~\cite{sgqn}       & Ours       \\ \midrule
\multirow{6}{*}{\rotatebox[]{270}{\textbf{Random Colors}}} & Walker,Walk         & 144{\footnotesize$\pm$19}  & 520{\footnotesize$\pm$91}  & 168{\footnotesize$\pm$90} & 400{\footnotesize$\pm$61}  & 468{\footnotesize$\pm$47} & 692{\footnotesize$\pm$68}  & 749{\footnotesize$\pm$61} & 823{\footnotesize$\pm$58} & 884{\footnotesize$\pm$20} & 785{\footnotesize$\pm$57} & \textbf{902{\footnotesize$\pm$46} }\\
                               & Walker,Stand        & 365{\footnotesize$\pm$79}  & 770{\footnotesize$\pm$71}  & 413{\footnotesize$\pm$61} & 644{\footnotesize$\pm$88}  & 797{\footnotesize$\pm$46} & 893{\footnotesize$\pm$12} & 933{\footnotesize$\pm$24} & 947{\footnotesize$\pm$26} & 960{\footnotesize$\pm$15} & 929{\footnotesize$\pm$12} & \textbf{972{\footnotesize$\pm$23}} \\
                               & Ball\_in\_cup,Catch & 151{\footnotesize$\pm$36}  & 365{\footnotesize$\pm$210} & 469{\footnotesize$\pm$99} & 541{\footnotesize$\pm$29}  & 563{\footnotesize$\pm$50} & 805{\footnotesize$\pm$28}  & 959{\footnotesize$\pm$5}  & 932{\footnotesize$\pm$32} & 964{\footnotesize$\pm$7}  & 864{\footnotesize$\pm$75} & \textbf{972{\footnotesize$\pm$10}} \\
                               & Finger,Spin         & 316{\footnotesize$\pm$119} & 402{\footnotesize$\pm$208} & 478{\footnotesize$\pm$46} & 667{\footnotesize$\pm$154} & 803{\footnotesize$\pm$72} & 793{\footnotesize$\pm$128} & 912{\footnotesize$\pm$6}  & 876{\footnotesize$\pm$45} & 922{\footnotesize$\pm$54} & 905{\footnotesize$\pm$43} & \textbf{928{\footnotesize$\pm$43}} \\
                               & Cartpole,Swingup    & 248{\footnotesize$\pm$24}  & 586{\footnotesize$\pm$52}  & 277{\footnotesize$\pm$80} & 590{\footnotesize$\pm$53}  & 630{\footnotesize$\pm$63} & 949{\footnotesize$\pm$19}  & 832{\footnotesize$\pm$23} & 760{\footnotesize$\pm$60} & 749{\footnotesize$\pm$46} & 840{\footnotesize$\pm$13} & \textbf{856{\footnotesize$\pm$40}} \\
                               & Cheetah,Run         & 76{\footnotesize$\pm$25}   & 100{\footnotesize$\pm$27}  & 109{\footnotesize$\pm$45} & 121{\footnotesize$\pm$79}  & 159{\footnotesize$\pm$28} & 228{\footnotesize$\pm$76}  & 273{\footnotesize$\pm$23} & 371{\footnotesize$\pm$51} & 369{\footnotesize$\pm$53} & 162{\footnotesize$\pm$38} & \textbf{375{\footnotesize$\pm$32} }\\ \bottomrule
\end{tabular}}
\end{center}

\caption{\textbf{Generalization on random colors environments.} Experiments are conducted on 6 challenging tasks in the DMC-GB. Our CG2A agent perform well over all tasks and exceeds the prior SOTA methods with a significant margin.}
\label{color}
\end{table*}

\subsection{Soft Gradient Surgery}

 Here, we propose to modify the standard average based gradient descent by incorporating a soft gradient surgery step before updating the neural parameters $\theta$ of agent policy.
Instead of directly computing similarity between pairwise gradient vectors, we focus on recognizing the element-level gradient conflicting component cross the all gradient items. Our approach aims to adjust the model parameters $\theta$ by modifying the gradient updates to point towards conflict-free direction and improve consistency across all data augmentation. Specifically, given a set of gradient vectors (one for each data augmentation), we construct semantic agreement gradients by retaining the components with the same sign and clipping the conflicting components with a damping factor $\gamma$ to restrain excessive exploration caused by conflicting components. In particular, we introduce the following element gradient mask $\vartheta$ as an indicator to determine which component is in conflict:
\begin{align}
\label{mask}
 \begin{split}
 \vartheta[j]= \left \{
\begin{array}{ll} 
    1,                 & \sum_{i=0}^{N} \text{sign}(g_i^j)= N+1, \\
    0,                                 & \text{otherwise},
\end{array}
\right. 
\end{split}
\end{align}
where $g_{i}^j$ denotes $j$-th component in the flattened gradient vectors $g_i$. 
Note that $\vartheta[\cdot] = 1$ denotes that such gradient component is consistently agree upon all data augmentations, so we preserve the complete gradient information. In contrast, $\vartheta[\cdot] = 0$ indicates the component is in conflict and would be clipped.
To precisely constrain the semantic-irrelevant information in conflicting gradient components, we introduce a damping factor $\gamma$, sampling from a uniform distribution $\gamma \in \textbf{U}(\alpha,\beta)$, to control the clipping ratio. Compared to constant clipping, random sampling can effectively enhance the policy exploration ability and prevent falling into local optima.
Hence, the gradient update procedure for gradient vector $g_i$ can be written as follows:
\begin{equation}
    \label{update}
    g_i = \vartheta \times g_i +  \gamma \times \sim \vartheta \times g_i, 
\end{equation}
where  $\sim\vartheta$ is obtained by applying the bitwise $NOT$ operator to the mask $\vartheta$.
The whole update procedure of the SGS algorithm is provided in Algorithm~\ref{sgs}. The computational overhead of our SGS is minimal, primarily involving simple operations such as sign and addition functions applied to the flatten gradient vectors.

\section{Experiments}
To evaluate the generalization performance and sample efficiency of our proposed CG2A, we compare it to several state-of-the-art methods on a set of standard tasks from the DMControl Generalization Benchmark (DMC-GB) and two vision based robotic manipulation tasks. 

\noindent\textbf{Setup.}
Following prior works~\cite{svea,sgqn}, we implement the SAC~\cite{sac} algorithm with random shift as baseline and adopt the same network architecture and hyper-parameter setup as Hansen~\etal\cite{soda} for all applicable methods. The observation for DMC-GB tasks is a sequence of three consecutive RGB frames with dimensions of $84\times84\times3$, except for robotic manipulation tasks which use a single frame. Besides, the hyper-parameter $\alpha$ and $\beta$ in SGS are set as 0.22 and 0.28 respectively.
In all experiments, the generalization evaluations are executed in a zero-shot paradigm and we report the average result over 5 times.

\noindent\textbf{Baselines and Data Augmentations.} 
To evaluate the generalization capability of our CG2A, we benchmark CG2A against strong baselines and several state-of-the-art methods: SAC~\cite{sac}, DrQ~\cite{drq}, DrQv2~\cite{drqv2}, RAD~\cite{rad}, PAD~\cite{pad} SODA~\cite{soda}, SVEA~\cite{svea}, TLDA~\cite{tlda}, PIE-G~\cite{pieg}, and SGQN~\cite{sgqn}. For all compared methods, we report the best performance in the available literature as well as in the reproduced results. Considering that most methods use data augmentation in one of their stages, we adopt random overlay~\cite{soda} as default, which mixup observations and random images from the Places365 dataset~\cite{places} $\mathcal{D}$, as follows:
\begin{equation}
    \tau_{overlay}(o) = (1-\mu) o + \epsilon, \epsilon \in \mathcal{D}_{place},
\end{equation}
where $\mu\in[0,1)$ is the interpolation coefficient and default set as 0.5. For $\mu$ values smaller than $0.20$, we consider the augmentation to be perceptually insensitive and label it as overlay-S for brevity. To reduce computational overhead, we choose three augmentations ($N=3$) to construct the augmentation combination, including random conv~\cite{rad}, random overlay~\cite{soda}, random overlay-S~\cite{soda}.
\subsection{Evaluation on DMC-GB}
The DMC-GB contains a set of vision-based continuous control tasks~\cite{dmc}, which allows agents to be trained in a fixed environment and evaluate generalization capability on unseen environments with distribution shifts, including \textit{random colors} and \textit{video backgrounds}.
For \textit{video backgrounds} setting, \textit{video easy} benchmark modifies solely the background of images with a distracting image, whereas the hard version extends this modification to include the ground and the shadows, which is more challenging. The training process includes 500,000 interaction steps with 4 action repeats as default, and the agents are evaluated with 100 episodes.

\noindent\textbf{Random Colors.}
The experimental results, as depicted in Table~\ref{color}, show that CG2A outperforms prior state-of-the-art methods in all tasks, indicating its superior performance. These results demonstrate that integrating augmentation combination can effectively enhance the robustness of agent to color change in unknown environment, which exposing the potential of augmentation combination mechanism for improving generalization in Visual RL.
\begin{table*}[]
\begin{center}
\renewcommand\arraystretch{1.1}
\resizebox{\textwidth}{!}{
\begin{tabular}{@{}ccccccccccccc@{}}
\toprule                    & DMCGB~\cite{soda}          & SAC~\cite{sac}        & DrQ~\cite{drq}         & DrQv2~\cite{drqv2}       & RAD~\cite{rad}         & PAD~\cite{pad}        & SODA~\cite{soda}        & SVEA~\cite{svea}        & TLDA~\cite{tlda}       & PIE-G~\cite{pieg}   & SGQN~\cite{sac}       & Ours       \\ \midrule
\multirow{6}{*}{\rotatebox[]{270}{\textbf{Video Easy}}} & Walker,Walk         & 245{\footnotesize$\pm$165} & 682{\footnotesize$\pm$8}  & 175{\footnotesize$\pm$117} & 608{\footnotesize$\pm$92}  & 717{\footnotesize$\pm$79} & 771{\footnotesize$\pm$66}  & 819{\footnotesize$\pm$71}  & 873{\footnotesize$\pm$83} & 871{\footnotesize$\pm$22}  & 910{\footnotesize$\pm$24} & \textbf{918{\footnotesize$\pm$20}} \\
                            & Walker,Stand        & 389{\footnotesize$\pm$131} & 873{\footnotesize$\pm$83}  & 560{\footnotesize$\pm$48}  & 879{\footnotesize$\pm$64}  & 935{\footnotesize$\pm$20} & 965{\footnotesize$\pm$7}   & 961{\footnotesize$\pm$8}   & 946{\footnotesize$\pm$6}  & 957{\footnotesize$\pm$12}  & 955{\footnotesize$\pm$9}  & \textbf{968{\footnotesize$\pm$6}}  \\
                            & Ball\_in\_cup,Catch & 192{\footnotesize$\pm$157} & 318{\footnotesize$\pm$157} & 453{\footnotesize$\pm$60}  & 363{\footnotesize$\pm$158} & 436{\footnotesize$\pm$55} & 939{\footnotesize$\pm$10}  & 871{\footnotesize$\pm$106} & 892{\footnotesize$\pm$68} & 922{\footnotesize$\pm$20}  & 950{\footnotesize$\pm$24} & \textbf{963{\footnotesize$\pm$28} }\\
                            & Finger,Spin         & 152{\footnotesize$\pm$8}   & 533{\footnotesize$\pm$119} & 456{\footnotesize$\pm$15}  & 334{\footnotesize$\pm$54}  & 691{\footnotesize$\pm$80} & 535{\footnotesize$\pm$52}  & 808{\footnotesize$\pm$23}  & 744{\footnotesize$\pm$18} & 837{\footnotesize$\pm$107} & 609{\footnotesize$\pm$61} & \textbf{912{\footnotesize$\pm$69}} \\
                            & Cartpole,Swingup    & 472{\footnotesize$\pm$26}  & 485{\footnotesize$\pm$105} & 267{\footnotesize$\pm$41}  & 391{\footnotesize$\pm$66}  & 521{\footnotesize$\pm$76} & 678{\footnotesize$\pm$120} & 702{\footnotesize$\pm$80}  & 671{\footnotesize$\pm$57} & 587{\footnotesize$\pm$61}  & 717{\footnotesize$\pm$35} & \textbf{788{\footnotesize$\pm$24}} \\
                            & Cheetah,Run         & 87{\footnotesize$\pm$21}   & 102{\footnotesize$\pm$30}  & 64{\footnotesize$\pm$22}   & 43{\footnotesize$\pm$21}   & 206{\footnotesize$\pm$34} & 184{\footnotesize$\pm$64}  & 249{\footnotesize$\pm$20}  & 308{\footnotesize$\pm$57} & 287{\footnotesize$\pm$20}  & 269{\footnotesize$\pm$33} & \textbf{314{\footnotesize$\pm$49}} \\ \midrule
\multirow{6}{*}{\rotatebox[]{270}{\textbf{Video Hard}}} & Walker,Walk         & 122{\footnotesize$\pm$47}  & 104{\footnotesize$\pm$22}  & 34{\footnotesize$\pm$11 }  & 80{\footnotesize$\pm$10}   & 189{\footnotesize$\pm$54} & 312{\footnotesize$\pm$32 } & 385{\footnotesize$\pm$63}  & 271{\footnotesize$\pm$55} & 600{\footnotesize$\pm$28}  & \textbf{739{\footnotesize$\pm$21}} & 687{\footnotesize$\pm$18} \\
                            & Walker,Stand        & 231{\footnotesize$\pm$57}  & 289{\footnotesize$\pm$49}  & 151{\footnotesize$\pm$13}  & 229{\footnotesize$\pm$45}  & 411{\footnotesize$\pm$36} & 736{\footnotesize$\pm$132} & 747{\footnotesize$\pm$43}  & 602{\footnotesize$\pm$51} & 852{\footnotesize$\pm$56}  & 851{\footnotesize$\pm$24} & \textbf{895{\footnotesize$\pm$35}} \\
                            & Ball\_in\_cup,Catch & 101{\footnotesize$\pm$37}  & 92{\footnotesize$\pm$23}   & 97{\footnotesize$\pm$27}   & 98{\footnotesize$\pm$40}   & 174{\footnotesize$\pm$71} & 381{\footnotesize$\pm$163} & 403{\footnotesize$\pm$174} & 257{\footnotesize$\pm$57} & 786{\footnotesize$\pm$47}  & 782{\footnotesize$\pm$57} & \textbf{806{\footnotesize$\pm$44}} \\
                            & Finger,Spin         & 25{\footnotesize$\pm$6}    & 71{\footnotesize$\pm$45}   & 21{\footnotesize$\pm$4}    & 15{\footnotesize$\pm$6}    & 144{\footnotesize$\pm$19} & 221{\footnotesize$\pm$48}  & 335{\footnotesize$\pm$58}  & 241{\footnotesize$\pm$29} & 762{\footnotesize$\pm$59}  & 540{\footnotesize$\pm$53} & \textbf{819{\footnotesize$\pm$38}} \\
                            & Cartpole,Swingup    & 153{\footnotesize$\pm$22}  & 138{\footnotesize$\pm$9}   & 130{\footnotesize$\pm$3}   & 117{\footnotesize$\pm$22}  & 255{\footnotesize$\pm$60} & 339{\footnotesize$\pm$87}  & 393{\footnotesize$\pm$45}  & 286{\footnotesize$\pm$47} & 401{\footnotesize$\pm$21}  & 428{\footnotesize$\pm$60} &\textbf{ 472{\footnotesize$\pm$24}} \\
                            & Cheetah,Run         & 28{\footnotesize$\pm$6}    & 32{\footnotesize$\pm$13}   & 23{\footnotesize$\pm$5}    & 21{\footnotesize$\pm$7}    & 35{\footnotesize$\pm$22}  & 94{\footnotesize$\pm$75}   & 105{\footnotesize$\pm$37}  & 90{\footnotesize$\pm$27}  & 134{\footnotesize$\pm$17}  & 144{\footnotesize$\pm$34} & \textbf{168{\footnotesize$\pm$16} }\\ \bottomrule
\end{tabular}}
\end{center}
\caption{\textbf{Generalization on video backgrounds environments.}
Episode return in two kind of dynamic video background environments, \eg, $\textit{video easy}$ (Top) and $\textit{video hard}$ (Bottom). Bold font indicates the best performance among all methods.
}
\label{video}
\end{table*}

\noindent\textbf{Video Background.}
As illustrated in Table~\ref{video}, our CG2A surpasses the baselines in 11 out of 12 instances in terms of mean cumulative rewards. Notably, CG2A achieves competitive performance with all prior methods in the context of video easy setting. In particular, for tasks such as ``Finger, Spin" and ``Cartpole, Swingup", CG2A obtains a substantial profit margin of 8.9\% and 10.1\% respectively, outperforming other state-of-the-art methods. Additionally, CG2A achieves near-perfect scores in such setting on the ``Walker, walk" and ``Ball in cup, catch" tasks while significantly reducing the empirical variance to an inconsequential level. Moreover, the agent trained with CG2A demonstrates robust policy acquisition in more challenging video hard environments. These results highlight the effectiveness of CG2A in enhancing the agents' performance and generalization ability in complex and dynamic scenarios.

\subsection{Evaluation on Robotic Manipulation Tasks}
To validate the performance of the agent in realistic scenarios, we follow prior work~\cite{svea,sgqn} and incorporate two goal-reaching robotic manipulation tasks, ``Reach" and ``Peg In Box", from the vision-based robotic manipulation simulator outlined in~\cite{look}. To provide a comprehensive view, the RGB camera is positioned in front of the entire setup, providing a third-person view with a large field of view encompassing the robot, target objects, and workspace. The ``Reach" is required to locate the goal with a red disc on the table and control the robotic gripper move to there. And ``Peg in Box" aim to guide the  robot insert a peg affixed to its arm into a box, which is more challenging. The position of the gripper and target objects is randomized in all tasks, and there are significant variations in lighting and texture between the training and testing environments. More training hyper-parameters details and environment descriptions are provided in Appendix.


We train all agents for 250, 000 steps with default setting and evaluate its generalization performance in comparison to the agents trained with SAC~\cite{sac}, SODA~\cite{soda}, SVEA~\cite{svea}, and SGQN~\cite{sgqn}. Table~\ref{robot_test} demonstrates that all agents trained with prior SOTA fail to maintain their performance when evaluated on the three test environments. Instead, our approach outperforms these baselines in all robotic manipulation tasks, achieving advanced mean cumulative rewards by a significant margin. Particularly, our approach achieves remarkable performance in the first environment (Test1), surpassing the previous methods by 99.3\% and 919.5\% respectively in the Reach and PegInBox tasks.

\begin{table}[]
\begin{center}
\renewcommand\arraystretch{1.2}
\resizebox{.47\textwidth}{!}{
\begin{tabular}{@{}ccccccc@{}}
\toprule
Task                            & Settings & SAC~\cite{sac}          & SODA~\cite{soda}         & SVEA~\cite{svea}         & SGQN~\cite{sgqn}       & Ours         \\ \midrule
\multirow{4}{*}{\rotatebox[]{270}{ \textbf{Reach}}}    & train    & 9.7{\footnotesize$\pm$2}    & 31.8{\footnotesize$\pm$1}   & 32.2{\footnotesize$\pm$5}   & 31.8{\footnotesize$\pm$1}  & \textbf{39.6{\footnotesize$\pm$4} }  \\
                                & test1    & -20.9{\footnotesize$\pm$16} & -30.9{\footnotesize$\pm$43} & -17.6{\footnotesize$\pm$10} & 14.4{\footnotesize$\pm$14} & \textbf{28.7{\footnotesize$\pm$1}}   \\
                                & test2    & -21.9{\footnotesize$\pm$14} & -20.2{\footnotesize$\pm$29} & -2.1{\footnotesize$\pm$39}  & 31.0{\footnotesize$\pm$3}  & \textbf{36.7{\footnotesize$\pm$4} }  \\
                                & test3    & -43.2{\footnotesize$\pm$6}  & -68.4{\footnotesize$\pm$30} & 1.4{\footnotesize$\pm$29}   & 29.2{\footnotesize$\pm$7}  & \textbf{35.4{\footnotesize$\pm$4}}   \\ \midrule
\multirow{4}{*}{\rotatebox[]{270}{\textbf{PegInBox}}} & train    & -46.7{\footnotesize$\pm$7}  & 180.1{\footnotesize$\pm$2}  & 177.5{\footnotesize$\pm$3 } & 183.9{\footnotesize$\pm$9} & \textbf{189.9{\footnotesize$\pm$11}} \\
                                & test1    & -20.9{\footnotesize$\pm$16} & 16.9{\footnotesize$\pm$44}  & -21.3{\footnotesize$\pm$10} & -72.0{\footnotesize$\pm$14}  & \textbf{155.4{\footnotesize$\pm$17} }\\
                                & test2    & -21.9{\footnotesize$\pm$14} & 0.7{\footnotesize$\pm$30}   & 96.8{\footnotesize$\pm$42}  & 110.7{\footnotesize$\pm$3} &\textbf{ 157.8{\footnotesize$\pm$22}} \\
                                & test3    & -43.2{\footnotesize$\pm$6}  & 73.6{\footnotesize$\pm$31}  & 40.5{\footnotesize$\pm$28}  & 154.6{\footnotesize$\pm$7} &\textbf{ 174.0{\footnotesize$\pm$21}}   \\ \bottomrule
\end{tabular}}
\end{center}
\caption{\textbf{Generalization on robotic manipulation tasks.} Our CG2A significantly outperforms other methods by a large margin in both tasks, with only a slight decrease in performance observed across all testing scenarios.}
\vspace{-10pt}
\label{robot_test}
\end{table}


\subsection{Sample Efficiency}

\begin{figure}
    \begin{center}
    \includegraphics[width=\linewidth]{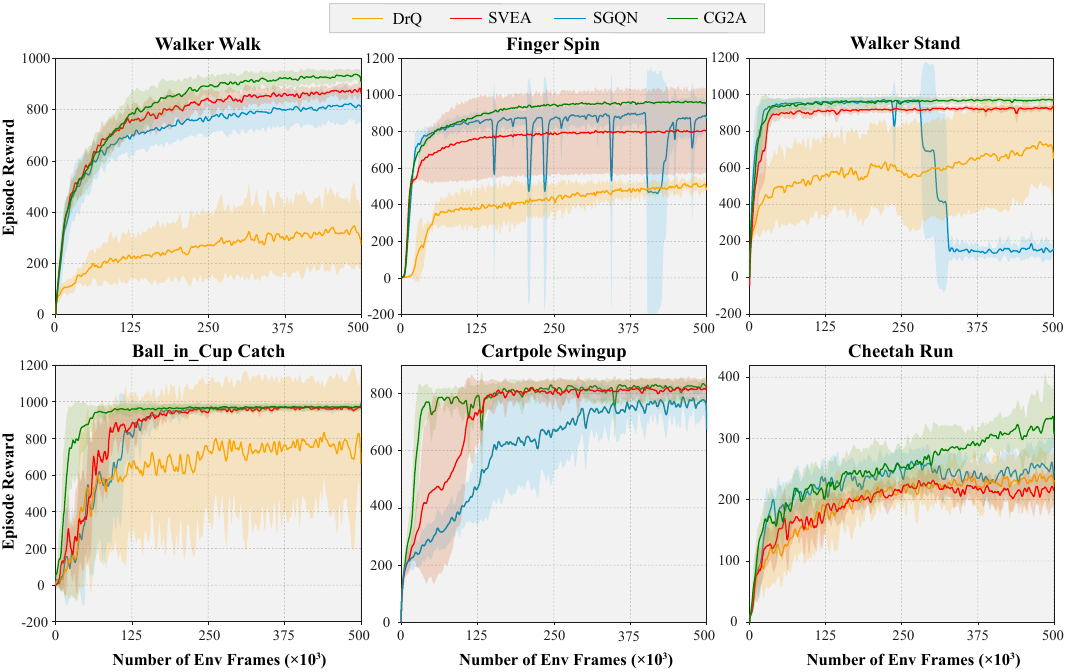}
    \end{center}
    \caption{\textbf{Training sample efficiency.} Comparison of CG2A (Green Line) with sample-efficient RL algorithms, including DrQ~\cite{drq} (Yellow Line), SVEA~\cite{svea} (Red Line) and SGQN~\cite{sgqn} (Blue Line). Our method achieve better performance on all tasks.}
    \vspace{-10pt}
    \label{curve_dmc}
\end{figure}

To verify the sample efficiency of our proposed CG2A, we compare our method with prior state-of-the-art methods, including DrQ~\cite{drq}, SVEA~\cite{svea}, and SGQN~\cite{sgqn}, on DMControl suite~\cite{dmc} and robotic benchmark. Figure~\ref{curve_dmc} demonstrates that our proposed CG2A agent significantly outperforms the other SOTA agents across selected tasks, in terms of asymptotic performance and sample efficiency, on all evaluation settings. 
Compared with SVEA~\cite{svea}, our method shows better performance and stability with a smaller variance. Notably, SGQN~\cite{sgqn} exhibits severe performance collapse on some tasks, which can be attributed to the limitations of saliency-based self-learning. Once the auxiliary task parameters get into a local dilemma, it can cause the performance of the RL agent to crash to maintain gradient balance. Therefore, our CG2A avoids the introduction of any learnable parameters for constructing the auxiliary task to ensure training stability. Besides, the CG2A converges to optimal performance for the `Ball\_in\_Cup, Catch' and `Cartpole, Swingup' tasks at least 100,000 training steps earlier than other methods like SVEA and SGQN. 
These experimental results demonstrate that the utilization of data augmentation combinations not only enhances the policy's generalization ability but also effectively improves sampling efficiency and training stability.

\subsection{Ablation Study}
To evaluate the effectiveness of our proposed CG2A, we conduct comprehensive ablation analyses to closely validate the individual components of the CG2A. All these agents are trained on four standard tasks from DMControl suite~\cite{dmc} with 500, 000 training steps and evaluated on the challenging $\textit{video hard}$ benchmark. More ablation results about additional tasks and various augmentation combinations are provided in Appendix.

\noindent\textbf{Effectiveness of Individual Components.} Compared with vanilla SAC algorithm, CG2A gathers multiple data augmentations to construct augmentation combination and enhance the SAC architecture with an adaptive weight solver GAS and a conflict-aware gradient fusion strategy named SGS. We perform an ablation study to investigate the effectiveness of individual components in CG2A and the results are shown in Table~\ref{ablation_c}. Individually, each of these features contributes significantly to the improvement of generalization performance across all environments. The introduction of augmentation combination provides great performance gains over vanilla SAC, which can achieve the comparable performance with SVEA~\cite{svea}. Notably, agents trained with naive augmentation combination lead to higher performance variances than the vanilla SAC agents. Heavy data augmentations lead to higher performance variances against the vanilla over all tasks, which also leads to the severe gradient conflicts. The SGS strategy provides the most significant performance gains and stabilize the training process through randomly clipping conflicting gradient components. The impact of hyper-parameter $\gamma$ in the SGS is also illustrate in the next section. Experimental results suggest that the agents trained with the complete CG2A achieve the most superior generalization performance on all tasks.

\begin{figure}
    \begin{center}
    \includegraphics[width=\linewidth]{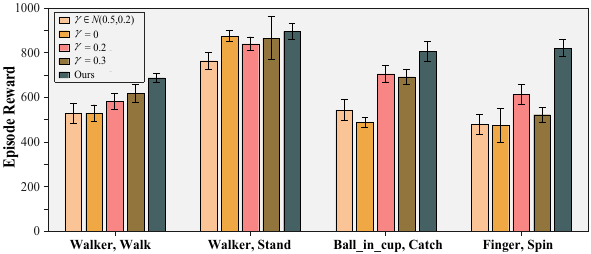}
    \end{center}
    \caption{Ablation study of the damping factor $\gamma$.} 
    \label{ablation_r}
    \vspace{-15pt}
\end{figure}

\noindent\textbf{Impact of Damping Factor $\gamma$.}
In our experiments, the value of the damping factor $\gamma$ was sampled from a uniform distribution with hyper-parameters $\alpha$ and $\beta$, which were obtained through grid search. To assess the sensitivity of the hyper-parameter $\gamma$ in SGS, we compared it with constant values of $\gamma\in{0, 0.2, 0.3}$ and other random distributions. Figure~\ref{ablation_r} shows that our method is robust to variations in the hyper-parameter $\gamma$ and achieves superior performance on most tasks. Notably, when $\gamma$ is set to 0, some gradient components are removed, limiting its performance upper bound.

\begin{table}[]
\begin{center}
\resizebox{0.99\linewidth}{!}{
\begin{tabular}{@{}ccc|cccc@{}}
\toprule
AC &
  GAS &
  SGS &
  \begin{tabular}[c]{@{}c@{}}Walker,\\ walk\end{tabular} &
  \begin{tabular}[c]{@{}c@{}}Walker,\\ stand\end{tabular} &
  \begin{tabular}[c]{@{}c@{}}ball\_in\_cup,\\ run\end{tabular} &
  \begin{tabular}[c]{@{}c@{}}Finger, \\ spin\end{tabular} \\ \midrule
  &   &   & 144 {\footnotesize$\pm$34} & 289 {\footnotesize$\pm$49} & 92 {\footnotesize$\pm$23}  & 71 {\footnotesize$\pm$45}  \\
\checkmark &   &   & 274 {\footnotesize$\pm$78} & 557 {\footnotesize$\pm$87} & 418 {\footnotesize$\pm$56} & 473 {\footnotesize$\pm$64} \\
\checkmark & \checkmark &   & 424 {\footnotesize$\pm$42} & 618 {\footnotesize$\pm$21} & 563 {\footnotesize$\pm$90} & 474 {\footnotesize$\pm$77} \\
\checkmark &   & \checkmark & 619 {\footnotesize$\pm$38} & 805 {\footnotesize$\pm$41} & 724 {\footnotesize$\pm$53} & 757 {\footnotesize$\pm$23} \\
\checkmark & \checkmark & \checkmark & \textbf{687 {\footnotesize$\pm$18}} & \textbf{895 {\footnotesize$\pm$35}} & \textbf{806 {\footnotesize$\pm$44}} & \textbf{819 {\footnotesize$\pm$38}} \\ \bottomrule
\end{tabular}}
\end{center}
\caption{Ablation study of individual components.}
\label{ablation_c}
\end{table}

\section{Conclusion}
In this paper, we integrate augmentation combination into visual RL to eliminate the generalization bias induced by single data augmentation, and propose Conflict-aware Gradient Agreement Augmentation, which can efficiently harmonize gradients with high-variance magnitudes and significantly mitigates performance degradation caused by gradient conflicts. Experimental results demonstrate that our method achieves state-of-the-art generalization performance with great sample efficiency. In the future, we will further explore the impact of the augmentation combination composition on generalization performance.
\section*{Acknowledgments}
This work was supported in part by Shanghai Municipal Science and Technology Major Project (No.2021SHZDZX0103), in part by Scientific Research Development Center in Higher Education Institutions by the Ministry of Education, China under Grant 2021ITA10013.

\clearpage
\normalem
\bibliographystyle{ieee_fullname}
\bibliography{egbib.bib}

\end{document}